# Gender-Based Comparative Study of Type 2 Diabetes Risk Factors in Kolkata, India: A Machine Learning Approach

Rahul Jain[1], Anoushka Saha[2], Gourav Daga[2], Durba Bhattacharya[2], Madhura Das Gupta[2], Sourav Chowdhury[2], Suparna Roychowdhury[2]

[1] Belle Vue Clinic, Kolkata
[2] St. Xavier's College (Autonomous), Kolkata

**Abstract:** Type 2 diabetes mellitus represents a prevalent and widespread global health concern, necessitating a comprehensive assessment of its risk factors. This study aimed towards learning whether there is any differential impact of age, Lifestyle, BMI and Waist to height ratio on the risk of Type 2 diabetes mellitus in males and females in Kolkata, West Bengal, India based on a sample observed from the out-patient consultation department of Belle Vue Clinic in Kolkata. Various machine learning models like Logistic Regression, Random Forest, and Support Vector Classifier, were used to predict the risk of diabetes, and performance was compared based on different predictors. Our findings indicate a significant age-related increase in risk of diabetes for both males and females. Although exercising and BMI was found to have significant impact on the risk of Type 2 diabetes in males, in females both turned out to be statistically insignificant. For both males and females, predictive models based on WhtR demonstrated superior performance in risk assessment compared to those based on BMI. This study sheds light on the gender-specific differences in the risk factors for Type 2 diabetes, offering valuable insights that can be used towards more targeted healthcare interventions and public health strategies.

**Keywords:** Machine Learning, Type2 Diabetes Mellitus, BMI, Waist to height ratio (Whtr), Logistic Regression, Random Forest, Support Vector Classifier.

## 1    Introduction

Type 2 diabetes is a rapidly growing non-communicable disease in India, ranking second worldwide with over 74 million cases in 2021. The IDF Diabetes Atlas (2021) [1] projects a 68% increase to 124.9 million individuals by 2045, posing a substantial health challenge.

Common Type 2 diabetes risk factors include excess weight, obesity, and sedentary lifestyle, but other factors still remain largely unexplored [2-4]. Existing anthropometric measures like BMI and waist circumference have limitations in assessing body fat distribution and accounting for ethnic variations [5]. Research suggests that waist-to-height ratio (Whtr), a measure of central obesity, may provide better insights into predicting cardio-metabolic abnormalities compared to BMI and waist circumference [6].

Global studies demonstrate variations in the performance of anthropometric measures for predicting Type 2 Diabetes Mellitus risk across diverse subpopulations [7]. Therefore, it is crucial to assess the predictive capabilities of these measures within distinct ethnic or geographic groups. Moreover, the prevalence of diabetes differs between obese men and women [7], with age consistently recognized as a significant risk factor for Type 2 diabetes [8].

This study is aimed towards exploring the application of Machine Learning techniques in healthcare, specifically within the context of predicting diabetes risk in Kolkata, West Bengal, India. The principal objective is to assess and compare the predictive capabilities of key variables: BMI, Whtr, age, and lifestyle, individually for males and females using appropriate Machine Learning algorithms. The analysis incorporates various machine learning algorithms, including logistic regression, support vector machine (SVM), and random forest classifier, following rigorous data preprocessing. The performance of these machine learning models is evaluated separately for males and females using receiver operating characteristic (ROC) analysis, considering metrics like the area under the curve (AUC) and Akaike information criterion (AIC) to assess model accuracy. Furthermore, the predictive outcomes of these Machine Learning algorithms are depicted through confusion matrices.

The rest of the paper is structured as follows: Section 2 presents a comprehensive description of the data and its collection procedure, including the definition of the variables under study. Section 3 briefly outlines the machine learning algorithms employed for this study. Section 4 is dedicated to a comprehensive analysis of the results of our work, and finally, in Section 5, the work is summarized, and concluding remarks are made.

## 2    Methods

A cross-sectional study was performed in the outpatient department of Belle Vue Clinic from March to May 2022, involving 428 patients. Among them, 211 were diagnosed with Type 2 diabetes mellitus, while 217 were tested negative. Following the guidelines of the Ethics Committee of the hospital, the participants were informed about the purpose of the study before collecting data.

Diabetes screening involved fasting plasma glucose measurement using the Hexokinase method after an overnight fast, with a cutoff of $\geq 126$ mg/dl indicating diabetes in the fasting state.

2 hour Post Prandial blood glucose of $\geq 200$mg/dl and/or HbA1c of $\geq 6.5$ (measured by HPLC method) or above were taken as Diabetic.

Height was measured with the subject standing using a wall-mounted scale, weight with a digital weighing machine while standing, and waist circumference at the midpoint of the Iliac crest anteriorly and the lower ribs with a flexible measuring tape, all while the subject stood and looked forward. Each measurement was taken twice, and the averages were recorded.

Information on Lifestyle activity and Gender was supplied by the respondents themselves. BMI and Waist to Height Ratio (henceforth referred to as Whtr) was derived from the variables described in Table1.

Considering the increased risk of diabetes with age, particularly at age ≥45, as established by previous studies [8] and the National Institute of Diabetes and Digestive and Kidney Diseases, the data was stratified by age and gender.

**Table 1.** Description of the variables under study

| Predictors | Type | Observed Range / Categories |
|---|---|---|
| Age | Continuous | Min: 20, Max: 88 |
| Sex | Categorical | 204 Males and 224 Females |
| Height | Continuous | Min:139 cm, Max: 189.5 cm |
| Weight | Continuous | Min: 41.8 kg, Max:144.4 kg |
| Waist Circumference | Continuous | Min=64; Max=143 |
| Lifestyle | Categorical | 3 categories: Does not exercise, Moderately exercises, Exercises daily. |

The interrelationship between various anthropometric variables was studied using correlation heat maps. The dataset was divided randomly into 80% training and 20% test sets. Multiple logistic regression, random forest, and support vector classifier models were applied separately for male and female Type 2 diabetes risk prediction. Model performance was evaluated using ROC analysis, with AUC and AIC considered for accuracy comparisons. R software was used to perform the statistical analysis.

## 3 Results

The mean age of the participants was 53.3 years, with standard deviation of 14.7 years. Mean height of the participants was 162.1 cm with a standard deviation of 9.6 cm. Mean weight of the participants was 72.7 kg with a standard deviation of 15.4 kg.

### 3.1 Data Visualization

In this section, the structure and interrelationships among the indicators are studied using correlation heat map and pair wise scatter plot.

The correlation heat map in Fig.1(a) shows high correlation (r = 0.82) between BMI and Whtr. Owing to the high level of multicollinearity between BMI and Whtr, separate predictive models were developed based on Age, Lifestyle, BMI and Age, Lifestyle and Whtr for males and females.

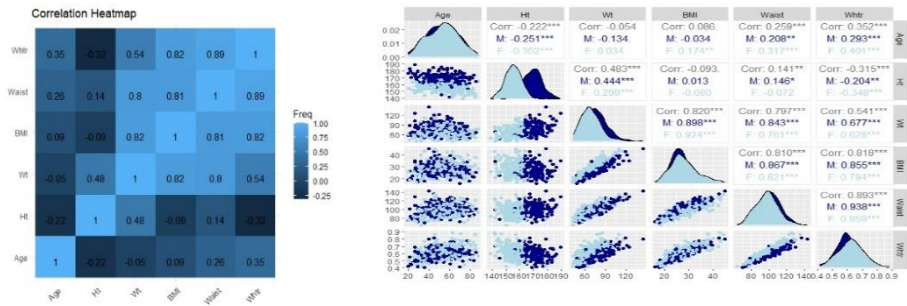

**Fig.(1a).** (Left Panel): Correlation heat map depicting the correlations between Age, BMI,Height, Weight, Waist circumference and Waist to height ratio. **Fig.(1b).** (Right Panel): Pairwise scatter plot for Age, Height, Weight, BMI, Waist circumference and Waist to height ratio.

The pair wise scatter plot in Fig.1(b), throws light on the correlation patterns between Age and weight, Age and BMI, height and BMI and Height and waist in males and females. For fixed levels of height or weight, on an average the females tend to have more BMI than the males. For fixed levels of weight, on an average the females tend to have higher waist to height ratio (Whtr).

### 3.2    Multiple Logistic Regression

Table 2 summarizes the results for the fitted logistic models based on Age, BMI and lifestyle for males and females. Age turns out to be a highly significant factor for both males and females. Surprisingly, BMI and Lifestyle turns out to be a significant factor only for males. As BMI has been detected as a high risk factor for Type 2 diabetes mellitus in different studies [10, 11, 12], further investigation is needed in this regard.

**Table 2.** Summary of the fitted regression models for Males and Females, based on Age, BMI and Lifestyle

| Predictors | Variable name | Estimates | | P value | | AUC | | AIC | |
|---|---|---|---|---|---|---|---|---|---|
| | | **Female** | **Male** | **Female** | **Male** | **Female** | **Male** | **Female** | **Male** |
| Intercept | Intercept | -1.980 | -4.175 | 0.028 | 0.001 | 0.701 | 0.810 | 223.26 | 188.29 |
| Age | Age(>=45) | 2.176 | 2.886 | $\approx 0$ | $\approx 0$ | | | | |
| BMI | BMI | 0.016 | 0.099 | 0.603 | 0.021 | | | | |
| Lifestyle | Exercises daily | 0.037 | -1.296 | 0.934 | 0.010 | | | | |
| | Moderately exercises | -0.174 | -0.982 | 0.646 | 0.030 | | | | |

The area under the ROC curve shows that BMI along with Age and Lifestyle has a good accuracy in predicting the risk of Type 2 diabetes mellitus for males. Among females it has a fair accuracy in predicting the risk of Type 2 diabetes mellitus. Thus we can observe that Age + BMI + Lifestyle performs better in predicting the risk of type 2 diabetes mellitus for males than for females. The AIC value for males is less than for females which again supports the fact that BMI along with Age and Lifestyle performs better for males than for females in predicting the risk of Type2 diabetes. See Fig.2 for diagrammatic summary of comparison of different predictive models using ROC curves.

Table 3 summarizes the results for the fitted predictive models based on Age, Waist to height ratio and lifestyle for males and females. Age and waist to height ratio turns out to be a highly significant factor for both males and females. Lifestyle is a significant factor for males only.

**Table 3.** Summary of the fitted regression models for Males and Females, based on Age, Whtr and Lifestyle

| Predictors | Variable name | Estimates | | P value | | AUC | | AIC | |
|---|---|---|---|---|---|---|---|---|---|
| | | **Female** | **Male** | **Female** | **Male** | **Female** | **Male** | **Female** | **Male** |
| Intercept | Intercept | -4.115 | -6.682 | 0.0021 | 0.0001 | 0.741 | 0.819 | 219.23 | 183.47 |
| Age | Age(>=45) | 1.954 | 2.586 | $\approx 0$ | $\approx 0$ | | | | |
| Whtr | Whtr | 4.341 | 9.069 | 0.0428 | 0.002 | | | | |
| Lifestyle | Exercises daily | 0.138 | -1.228 | 0.7647 | 0.016 | | | | |
| | Moderately exercises | -0.187 | -1.048 | 0.6235 | 0.023 | | | | |

The area under the ROC curve shows that Whtr along with Age and Lifestyle has a good accuracy in predicting the risk of Type 2 diabetes mellitus for males. Among females it has a fair accuracy in predicting the risk of type 2 diabetes mellitus. Thus we can observe that Whtr+Age+Lifestyle performs better in predicting the risk of Type 2 diabetes mellitus for males than for females. The AIC value for males is less than the AIC value for females, which again supports the fact that Whtr along with Age and Lifestyle performs better for males than for females in predicting the risk of Type 2 Diabetes.

From the AUC and AIC values, we also see that Age+Whtr+Lifestyle performs better than Age + BMI + Lifestyle for both males and females in predicting the risk of type2 diabetes mellitus. Fig.2 summarizes the performance of various predictive models based on ROC curves.

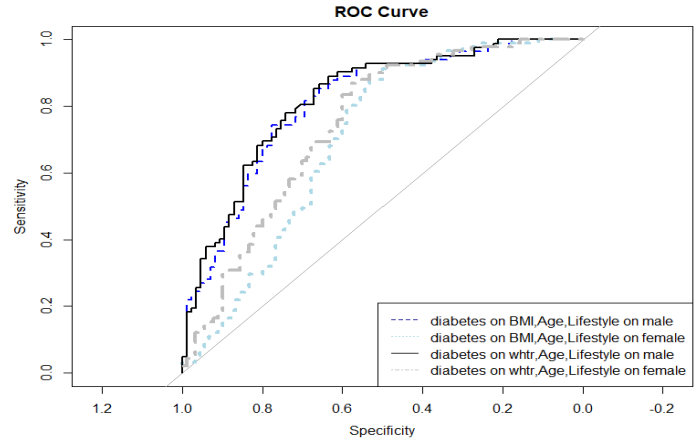

**Fig.2.** Pair wise scatter plot for Age, Height, Weight, BMI, Waist circumference and Waist to height ratio.

We also note that the estimated odds of Type 2 diabetes mellitus increases multiplicatively by 1.104 for one unit increase in BMI in males controlling for other variables i.e. there is a 10.4% increase. However, the estimated odds of Type 2 diabetes mellitus increases multiplicatively by 8686.27 for one unit increase in Whtr in males controlling for other variables. For females we see that the estimated odds increases multiplicatively by 76.85 for one unit increase in Whtr controlling for other variables.

On controlling for BMI and Lifestyle, the estimated odds of Type 2 diabetes mellitus for a male having Age ≥45 is 17.92 times of the estimated odds for a male having age<45.

On controlling for Whtr and Lifestyle, the estimated odds of Type 2 diabetes mellitus for a male having Age ≥45 is 13.27 times of the estimated odds for a male having age.

In case of Lifestyle, we see that controlling Age and BMI the estimated odds of Type 2 diabetes mellitus is in Males who Moderately Exercises and Exercises Daily are respectively 0.3734 times and 0.2736 times the odds for Males who do not exercise. In case of Lifestyle, we see that controlling Age and Whtr the estimated odds of Type 2 diabetes mellitus is in Males who Moderately Exercises and Exercises Daily are respectively 0.3506 times and 0.2928 times the odds for Males who do not exercise.

The confusion matrix obtained from predictions on the test set was used to calculate the performance in Table 4 and Table 5.

**Table 4.** Confusion matrix and performance measures for Logistic Regression model based on Age, BMI and Lifestyle

| Females | | | Males | | |
|---|---|---|---|---|---|
| | **Predicted** | | | **Predicted** | |
| **Observed** | **0** | **1** | **Observed** | **0** | **1** |
| **0** | 10 | 2 | **0** | 9 | 6 |
| **1** | 15 | 16 | **1** | 8 | 14 |
| Accuracy = 0.604 | | | Accuracy = 0.621 | | |
| Recall = 0.400 | | | Recall = 0.529 | | |
| Precision = 0.833 | | | Precision = 0.600 | | |
| F1 Score = 0.541 | | | F1 Score = 0.562 | | |

**Table 5.** Confusion matrix and performance measures for Logistic Regression model based on based on Age, Whtr and Lifestyle

| Females | | | Males | | |
|---|---|---|---|---|---|
| | **Predicted** | | | **Predicted** | |
| **Observed** | **0** | **1** | **Observed** | **0** | **1** |
| **0** | 11 | 2 | **0** | 11 | 8 |
| **1** | 14 | 16 | **1** | 6 | 12 |
| Accuracy = 0.628 | | | Accuracy = 0.622 | | |
| Recall = 0.440 | | | Recall = 0.647 | | |
| Precision = 0.846 | | | Precision = 0.579 | | |
| F1 Score = 0.579 | | | F1 Score = 0.611 | | |

### 3.3 Random Forest

Random forest models were fitted, separately for males and females, considering the predictors Age, Lifestyle, BMI and Whtr. We have taken both BMI and Whtr in the same model because random forest algorithm is not much affected by the presence of multicollinearity. The number of trees constructed was 100 and the number of variables considered at each split was 2. The out-of-bag estimate of error rate is 32.34%. The confusion matrices (Table 6) obtained from predictions on the test set were used to evaluate the performance measures for the different models.

**Table 6.** Confusion matrix and performance measures for Random Forest model based on Age, BMI, Whtr and Lifestyle

| Females | | | Males | | |
|---|---|---|---|---|---|
| | Predicted | | | Predicted | |
| Observed | 0 | 1 | Observed | 0 | 1 |
| **0** | 14 | 3 | **0** | 11 | 6 |
| **1** | 11 | 15 | **1** | 6 | 14 |
| Accuracy = 0.674 | | | Accuracy = 0.676 | | |
| Recall = 0.560 | | | Recall = 0.647 | | |
| Precision = 0.824 | | | Precision = 0.647 | | |
| F1 Score = 0.667 | | | F1 Score = 0.647 | | |

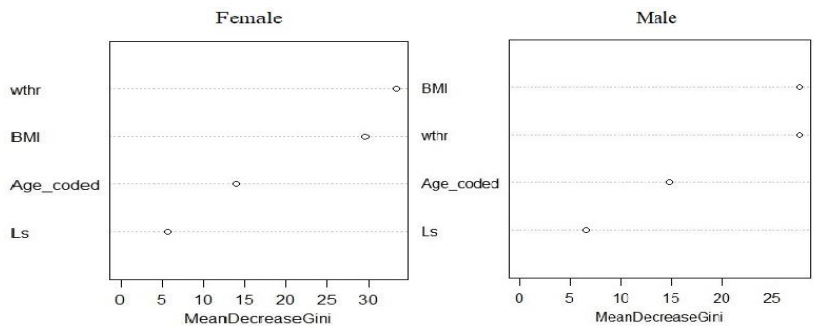

**Fig.3.** Variable Importance Plot for the Random Forest Model based on Age, BMI, Whtr and Lifestyle for Females and Males

The variable importance plot obtained from the random forest model for females and males separately in Fig.3 shows that the mean decrease in Gini index is the maximum for Whtr followed by that of BMI. This implies that the total decrease in the node impurity that results from split over Whtr is the maximum as compared to splits over the other variables. For males, the mean decrease in Gini index is almost the same for Wthr and BMI, in Fig.3. Hence, we can say that Whtr is the most important predictor for Type 2 Diabetes Mellitus in both males and females. This conclusion is at par with the one drawn from the Logistic Regression models.

### 3.4    Support Vector Classifier

The Support Vector Classifier algorithm is also not affected by the presence of multicollinearity. Thus, separate models for males and females were fitted, considering the predictors Age, Lifestyle, BMI and Whtr. The minimum error was obtained for cost C=0.1 using 10-fold cross validation.

The confusion matrices (Table 7) for the models are used to then evaluate the performance measures based on the test data.

**Table 7.** Confusion matrix and performance measures for Random Forest model based on Age, BMI, Whtr and Lifestyle

| Females | | | Males | | |
|---|---|---|---|---|---|
| | **Predicted** | | | **Predicted** | |
| **Observed** | **0** | **1** | **Observed** | **0** | **1** |
| **0** | 10 | 2 | **0** | 9 | 3 |
| **1** | 15 | 16 | **1** | 8 | 17 |
| Accuracy = 0.605 Recall = 0.400 Precision = 0.833 F1 Score = 0.551 | | | Accuracy = 0.703 Recall = 0.530 Precision = 0.750 F1 Score = 0.621 | | |

## 4 Discussion and Conclusion

In conclusion, our gender based comparative analysis of machine learning algorithms highlighted Random Forest's superior accuracy and F1 score, recommending it as the preferred model for diabetes risk prediction. While Logistic Regression showed consistent accuracy, precision favored females. Support Vector Classifier excelled for males but lagged for females. Our findings support Random Forest as the optimal choice in this case-controlled cross-sectional study in Kolkata for diabetes risk assessment.

A noteworthy finding in our study is the insignificance of BMI in predicting diabetes risk for females, contrary to previous research [2-4]. Additional investigation via the Mann-Whitney U test yielded a p-value of 1, indicating no significant difference in BMI medians between diabetic and non-diabetic females.

Our study identifies Waist-to-Height Ratio and age ≥45 as significant risk factors for Type 2 diabetes in both genders in Kolkata, emphasizing the need for increased preventive measures beyond this age. While exercise showed a protective effect in males, it was insignificant in females, warranting further investigation with a larger sample size.

While BMI and lifestyle are significant factors for Type 2 diabetes in Kolkata males, they lack significance in females. Importantly, Whtr proves to be a superior predictor for both genders, highlighting its significance for diabetes risk assessment. These findings emphasize the need for thoughtful model selection in gender-specific predictive modeling.

In future we plan to incorporate more anthropometric variables to study the risk of diabetes more comprehensively based on a larger data set under a Bayesian framework.